# SATELLITE CAPTIONING: LARGE LANGUAGE MODELS TO AUGMENT LABELING

Grant Rosario[1] and David Noever[2]

PeopleTec, 4901-D Corporate Drive, Huntsville, AL, USA, 35805
[1]grant.rosario@peopletec.com    [2]david.noever@peopletec.com

## ABSTRACT

*With the growing capabilities of modern object detection networks and datasets to train them, it has gotten more straightforward and, importantly, less laborious to get up and running with a model that is quite adept at detecting any number of various objects. However, while image datasets for object detection have grown and continue to proliferate (the current most extensive public set, ImageNet, contains over 14m images with over 14m instances), the same cannot be said for textual caption datasets. While they have certainly been growing in recent years, caption datasets present a much more difficult challenge due to language differences, grammar, and the time it takes for humans to generate them. Current datasets have certainly provided many instances to work with, but it becomes problematic when a captioner may have a more limited vocabulary, one may not be adequately fluent in the language, or there are simple grammatical mistakes. These difficulties are increased when the images get more specific, such as remote sensing images. This paper aims to address this issue of potential information and communication shortcomings in caption datasets. To provide a more precise analysis, we specify our domain of images to be remote sensing images in the RSICD dataset and experiment with the captions provided here. Our findings indicate that ChatGPT grammar correction is a simple and effective way to increase the performance accuracy of caption models by making data captions more diverse and grammatically correct.*

## 1. INTRODUCTION

Image captioning is becoming a prevalent multimodal AI task, aided by the rise of image translation, object detection, semantic segmentation, etc. [1, 2, 3]. In addition, recent research in Natural Language Processing (NLP) has provided new methods and models for annotating images with captions [4]. Accurately captioning a photo could allow one to build a large dataset of remote-sensing geographical pictures and a model to retrieve specific contextual scenes based on a text description [5]. However, some of the most used and state-of-the-art (SOTA) general image caption datasets today are based on manual annotations prone to numerous errors [6, 7, 8]. These errors can be obvious, such as incorrect grammar or spelling, or more subtle, such as overusing a keyword or descriptor. In addition, as an image becomes more abstract or less specific, such as remote sensing images, the image can become more challenging to describe in a short-form caption. While recent research has proposed solutions such as correcting, pruning, or augmenting the caption vocabulary [9], modern large language models (LLMs) provide a straightforward and efficient way of correcting errors in captions.

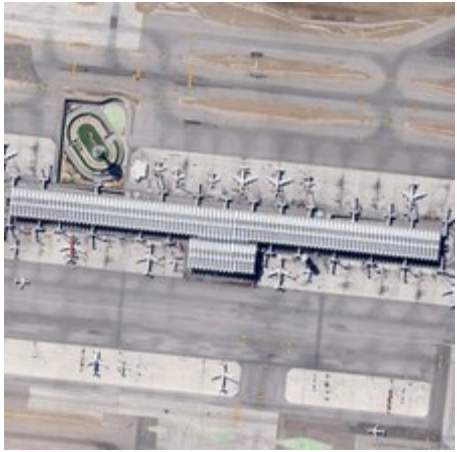

'raw': many planes are parked next to a long building in an airport .
'sentid': 0

'raw': many planes are parked next to a long building in an airport .
'sentid': 1

'raw': many planes are parked next to a long building in an airport .
'sentid': 3

'raw': many planes are parked next to a long building in an airport .
'sentid': 4

'raw': many planes are parked next to a long building in an airport .
'sentid': 5

**Figure 1: Original RSCID database caption. Note the five captions are identical.**

We approach this problem with the simple idea that current SOTA LLMs, such as ChatGPT, are extremely capable of generating human-like responses; therefore, when prompted appropriately, ChatGPT could read a sentence and potentially repair grammatical mistakes, remove unnecessary words, or increase word diversity. Figure 1 provides an example of one of the issues we aim to resolve: word repetition. The five captions in the Remote Sensing Image Captioning Dataset (RSICD) dataset [10] for this image are the same, which could cause overfitting issues and undesired pattern recognition for these images. Since different remote sensing images can contain high crossover with minor differences (similar colors, city layouts, road patterns), we believe the high repetition of vague descriptions results in a less accurate descriptor model. To highlight the capability of LLMs, Table 1 shows a comparison of 2 images from the RSICD dataset and their 5 human annotated captions vs a comprehensive caption generated by ChatGPT. We can observe that ChatGPT is plenty capable of generating detailed captions for a remote sensing image.

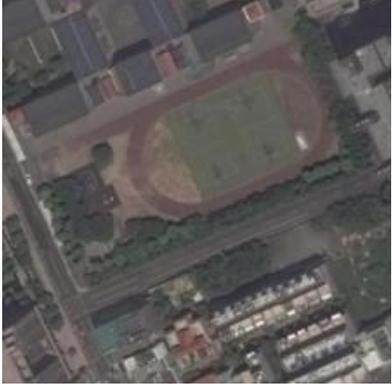

| Example Image | Human Caption Labels | LLM Caption Label |
| --- | --- | --- |
|  | • this halfold playground is wreathed in lush plants and buildings<br>• two small basketball fields and straight road beside<br>• many buildings and some green plants are around playground<br>• playground is surrounded by many buildings and two roads<br>• playground next to road is surrounded by many trees and houses | Educational Athletics: The image displays an athletic track commonly associated with a school or college, positioned between academic buildings to the north and commercial structures to the south, highlighting its role as a central feature of an educational campus in an urban setting. |

| 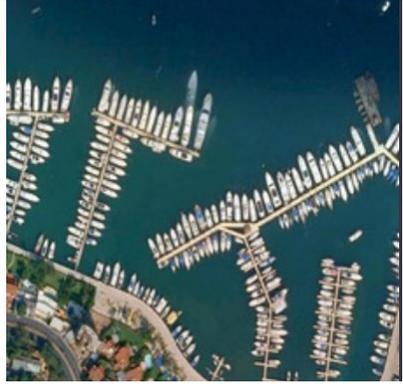 | many boats are in port near some buildings | Marina Aerial View: An organized array of boats and yachts docked in a marina, with the jetties extending like branches into the blue waters, captured from an overhead perspective |

Table 1: A table showing images from the RSICD dataset (right), the 5 captions per image within the dataset (middle) and new captions completely generated from ChatGPT (right)

| 'prompt': *"You are a helpful assistant that follows instructions extremely well. The following sentence may have several grammatical errors, please respond with a grammatically correct sentence that means the same thing and if possible, is more concise."* |
|---|
| 'original caption: *"many planes are parked next to a long building in an airport ."* |
| 'GPT-corrected caption: *"Several planes are parked alongside a lengthy building at the airport."* |

Table 2: (top) Our input prompt to ChaptGPT. (middle) The corresponding image caption of the original RSICD dataset. (bottom) The corresponding image caption of our grammar-corrected RSICD dataset

## 2. METHODS

The plan for this research ultimately includes comparing the results of a remote sensing image captioning model trained from the original RSICD dataset vs. an augmented RSICD dataset, which modifies the original captions based on the results from the "get-3.5-turbo" model of ChatGPT given our prompt. Table 2 shows the prompt we fed into the model, the resulting sentence, and the original RSCID sentence. While our prompt focuses on correcting grammatical errors and brevity, the API for ChatGPT also includes a 'temperature' variable used to increase the randomness of the responses, resulting in increased textual depth and avoiding high word repetition [11]. We left the variable at its default value of 1.0, which we believe provided appropriate vocabulary uniqueness while still providing sensible descriptions.

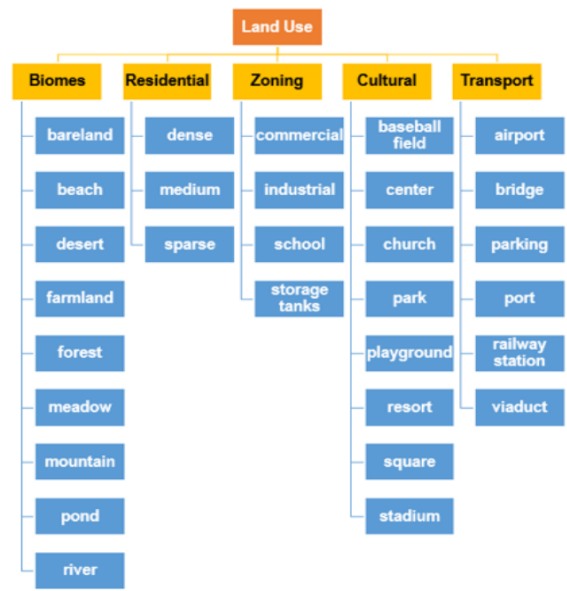

**Figure 2:** RSICD Scene classes

## 2.1. Dataset Descriptions

Each image in the RSICD dataset derives from satellite imaging services such as Google or Baidu at varying resolutions and sizes of 224 x 224 pixels. Each overhead photo is associated with five human-annotated captions, most duplicates. Each caption adheres to describing the image in the context of it belonging to one of 30 unique scene classes, shown in Figure 2.

As for the text in RSICD, the description diversity consists of 2643 unique words and 50,000 descriptions written by human annotators. However, over 60% of the descriptions are duplicates like those in Figure 1. Furthermore, approximately 14% of the text includes misspellings, broken syntax, and incorrect grammar.

|  | **RSICD Original** | **GPT-Corrected RSICD** |
|---|---|---|
| **Word Diversity** | 2643 unique words | 5365 unique words |
| **Rare One-time Words** | 925 rare words | 2102 rare words |
| **Ratio of Misspelled Words** | 1.04% | 0.7% |

**Table 3:** Showing the data differences between the original RSICD and our GPT-Augmented RSICD dataset based on word diversity, one-time word count, and misspelling ratio.

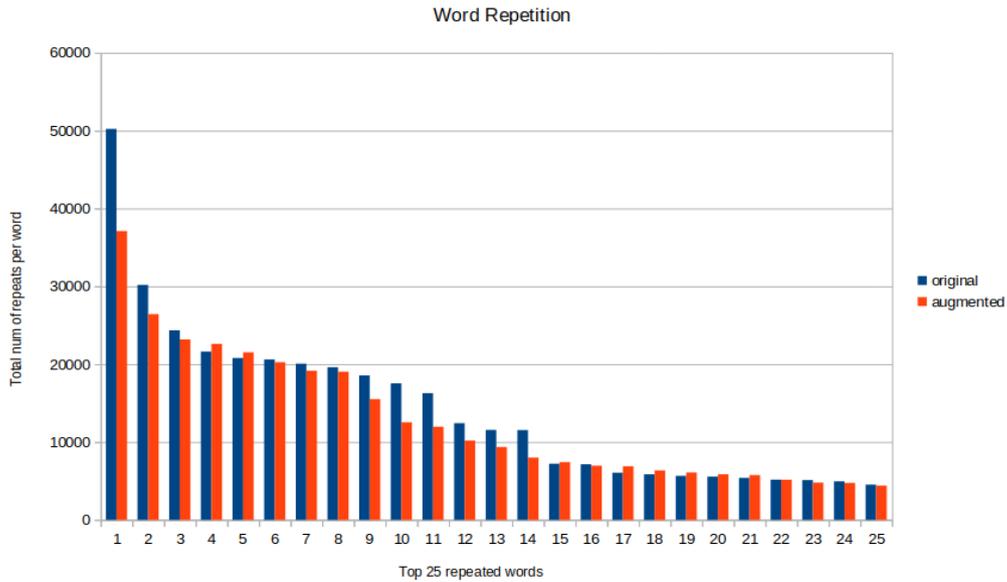

**Figure 3:** Original RSCID vs GPT-Corrected RSCID: word repetition

A prompt request to ChatGPT quickly resolves these issues, which could return a corrected sentence to replace the original. Table 3 provides some metrics of the two datasets we used: the original RSICD and the augmented RSCID with repaired sentences based on the original data. We can observe a higher diversity of words used, and a lower ratio of misspelled words in our augmented dataset, however it is interesting that we have a significantly higher number of one-time rare words in our new dataset which could be considered a negative since it makes it more difficult to recognize textual patterns in accordance with the image. Figure 3 shows a graph of the two datasets based on word repetition. Here, we can observe that our new dataset repeats words fewer times than the original which we believe should lead to more natural sounding captions.

## 2.2. Image Models Used for Feature Extraction

We wanted to be sure the models we used had varying architecture sizes and structures. The initial model used was ResNet-101 due to its high accuracy, although it is large and complex [12]. We then applied VGG-16 and VGG-19 due to their simplicity, although they can also be considered significant due to the multiple layers needed for accurate classifications [13]. Lastly, we applied MnasNet, which is also a convolution network but, unlike the others, is designed for mobile devices, so although it may not be as accurate as ResNet-101, it still achieves 75.2% Top1 accuracy while being able to run with 78ms latency on a mobile device [14].

## 2.3. Training

Our method for training consisted of following the encoder-decoder pipeline based on Xu et al.'s Show, Attend, and Tell caption generation paper. We begin our training approach by implementing transfer learning, using a pre-trained model of either ResNet-101, VGG-16, VGG-19, or MnasNet. These networks make up our encoder, which will encode the image into a concentrated shape of 14x14 with a specific number of channels depending on the network architecture, and since it's pretrained, it will be detecting objects in the image. We then pass the encoded image to a decoder, which generates a caption based on the image. Since we're dealing with text, we use a recurrent neural network with long short-term memory (LSTM) [15]. We also implement the Attention method in our decoder, which allows observing different parts of the image at various points during the training sequence by computing the pixel weights and using previously generated weight sequences to estimate the importance of the next pieces of the sequence [16].

|            | Original RSICD | GPT-Corrected RSICD |
|------------|----------------|---------------------|
| **ResNet-101** | 0.6859 | 0.7033 |
| **VGG-16** | 0.6853 | 0.6863 |
| **VGG-19** | 0.6579 | 0.6631 |
| **MnasNet** | 0.6312 | 0.6846 |

**Table 4:** Showing the final validation METEOR scores for both the original and GPT-corrected dataset for each feature extraction algorithm.

## 3. RESULTS

Since we are generating new captions that increased word diversity, we used the METEOR score to compare the results between the original RSICD dataset and our GPT-augmented dataset [17]. This was due to the increased diversity of our output captions from ChatGPT being likely to result in a worse BLEU score, whereas METEOR seems more resistant to diverse references. Table 4 shows the results of the validation METEOR scores for each dataset. We can see that while some are less statistically significant than others, there is a general improvement in caption readability and naturalism when we use our GPT-corrected RSICD dataset, supporting the notion that LLMs can provide improved performance for caption generation if prompted to correct any grammatical errors within the given dataset. Figure 4 shows the two datasets token-by-token results of the attention-based captioning algorithm using ResNet-101 as a backbone. One can observe that in this instance, the GPT-corrected dataset was able to train a more detailed, informative, and natural sounding caption compared to the original dataset.

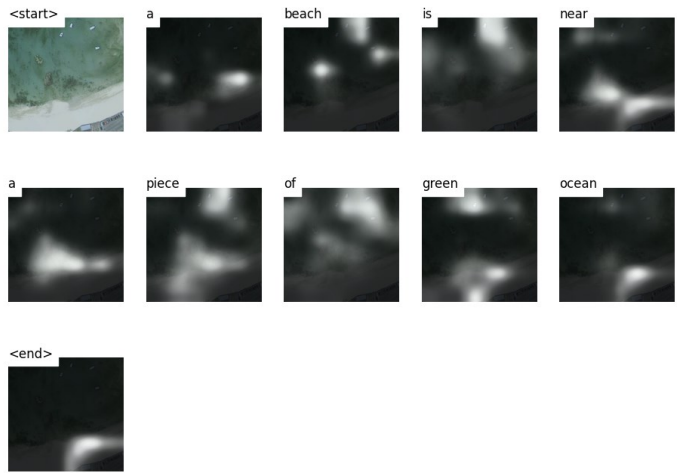

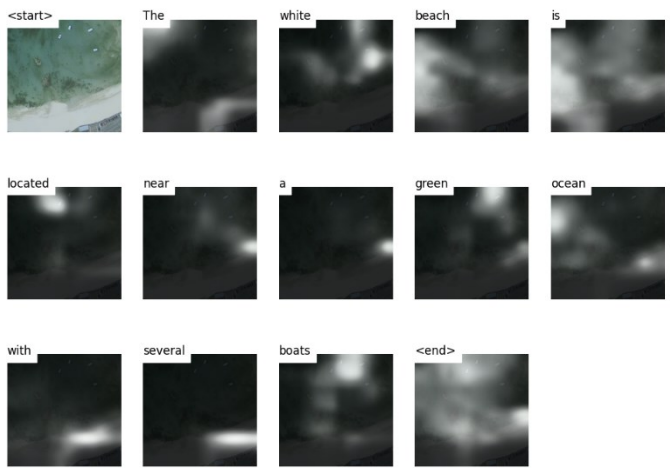

**Figure 4:** Token-by-token caption generation of ResNet-101 based model. (top): Original RSICD dataset. (bottom): GPT-corrected dataset.

## 4. DISCUSSION

This research has tested the ability for a LLM to be used as a dataset enhancer for a ML captioning model. The models were trained on the original RSCID dataset first, then the RSCID dataset captions were run through a custom pipeline using a LLM to correct language and grammar mistakes, decrease vagueness, and increase caption detail. The models used to compare the two datasets ranged from very large (ResNet) to very small (MnasNet) in order to show that enhancements could potentially be made across all model complexities.

### 4.1 Human-labeling Symptoms

While human-labeled data is beneficial for providing natural sounding language, the monotony and challenge of labelling such large amounts of images brings it's own issues. Word repetition can easily cause a model to overfit certain labels thus resulting in less generalization while too much word diversity can cause a model to have a tougher time training on certain scenes. Using a diverse enough vocabulary, avoiding repetition, and also avoiding extremely rare words can be challenging for humans due to unique knowledge and ellequency across labelers. Not to mention language barriers for people who are labelling in a non-dominant language. Table 3 and Figure 3 highlight these issues which we can observe are singifcantly alleviated with the use of LLMs.

### 4.2 Further Research Opportunities

Large Language Models are evolving rapidly and gaining new features nearly every month. During this experiment, OpenAI added several image translation features such as their image generator tool, DALLE, to GPT, enabled more customization options for models, and other competitors such as Google Bard have added photo and video features to increase their models knowledge base [18, 19]. Due to our findings and the ever increasing capabilities and advancements in LLMs, we believe these tools will eventually be a standard enhancement for any sort of ML modelling experiment.

Further research in this area could expand upon these findings by adding complexity to the model such as different language filters, pure synthetic caption vs human caption, or more advanced feature networks such as visual transformers.

## ACKNOWLEDGMENTS

The authors thank the PeopleTec Technical Fellows program for its encouragement and project assistance.

**Authors**

| | |
|---|---|
| **Grant Rosario** has research experience in embedded applications and autonomous driving applications. He received his Masters from Florida Atlantic University in Computer Science and his Bachelors from Florida Gulf Coast University in Psychology. | 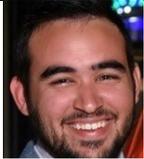 |
| **David Noever** has research experience with NASA and the Department of Defense in machine learning and data mining. He received his BS from Princeton University and his Ph.D. from Oxford University, as a Rhodes Scholar, in theoretical physics. | 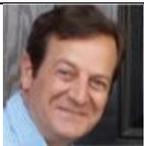 |